\documentclass[runningheads,a4paper]{llncs}

\usepackage{amssymb}
\usepackage{graphicx}
\usepackage{algorithm}
\usepackage[noend]{algpseudocode}

\usepackage{placeins} 
\usepackage{flafter}  

\usepackage{url}
\urldef{\mailsa}\path|{takoua.kefi, medbecha.kaaniche, adel.bouhoula}@supcom.tn| 
\urldef{\mailsc}\path|{ksantini}@uwindsor.ca|    
\newcommand{\keywords}[1]{\par\addvspace\baselineskip
\noindent\keywordname\enspace\ignorespaces#1}

\usepackage{amsmath}
\usepackage{epsfig}

\begin{document}

\mainmatter  

\mainmatter  

\title{Incremental One-Class Models for Data Classification}

\titlerunning{Incremental One-Class Models for Data Classification}
%
%
\author{Takoua Kefi$^{1}$  %
\and Riadh Ksantini$^{1,2}$\and Mohamed B\'{e}cha Ka\^{a}niche$^{1}$\and\\
 Adel Bouhoula$^{1}$}
\authorrunning{Kefi T.\and Ksantini R.\and Ka\^{a}niche M.B.\and Bouhoula A.}

\institute{$^{1}$Research Unit: S\'{e}curit\'{e} Num\'{e}rique ,\\
Higher School of Communication of Tunis, Tunisia\\
\mailsa\\
$^{2}$University of Windsor, 401, Sunset Avenue, Windsor, ON, Canada\\
\mailsc\\}

%
%

\maketitle

\begin{abstract} In this paper we outline a PhD research plan. This research contributes to the field of one-class incremental learning and classification in case of non-stationary environments.  The goal of this PhD is to define a new classification framework able to deal with very small learning dataset at the beginning of the process and with abilities to adjust itself according to the variability of the incoming data which create large scale datasets. As a preliminary work, incremental Covariance-guided One-Class Support Vector Machine is proposed to deal with sequentially obtained data. It is inspired from COSVM which put more emphasis on the low variance directions while keeping the basic formulation of incremental One-Class Support Vector Machine untouched. The incremental procedure is introduced by controlling the possible changes of support vectors after the addition of new data points, thanks to the Karush-Kuhn-Tucker conditions, that have to be maintained on all previously acquired data. Comparative experimental results with contemporary incremental and non-incremental one-class classifiers on numerous artificial and real data sets show that our method results in significantly better classification performance.
\end{abstract}

\section{Motivation and Description of the Problem Area}

Supervised learning is the task of finding a function which relates inputs and targets. The objective is to learn a model of  dependency of the  targets on the inputs. The ultimate goal is to be able to make accurate predictions $t$ for unseen values of $x$. Typically, we base our predictions upon some function defined over the input/training space, and learning is the  process of inferring the parameters of this function. A new representation of data is necessary to learn nonlinear relations with a linear classifier. This is equivalent to applying a fixed nonlinear mapping of the data to a feature space, in which the linear classifier can be used. 
In real world situations, such as, measurements collected from a nuclear power plant or rare medical diseases, it is usually assumed that information from normal operation are easy to collect during a training process, but most faults do not, or rarely appear. Therefore, One-Class Classification (OCC) \cite{OCC} should be used to detect abnormal or faulty behavior. 
To solve OCC problems, several methods have been proposed and different concrete models have been constructed. All these techniques can be divided into three main approaches \cite{ConceptLearning}: density-based methods, boundary-based methods and reconstruction-based methods.

The key limitation of the existing categories of OCC methods is that none of them consider the full scale of information available. For instance, using a density based method, we only focus on high density area and neglect areas with lower training data density, although they represent valid targets. In boundary-based methods, since only boundary data points are considered to build the model, the overall class is not completely considered. Moreover, in \cite{featpca}, it has been shown that projecting the data in the high variance directions (like PCA) will result in higher error (bias), while retaining the low variance directions will lower the total error. As a solution, Covariance Guided One-Class Support Vector Machine (COSVM) \cite{COSVM} has been proposed to put more emphasis on the low variance directions while keeping the basic formulation of incremental OSVM untouched. The covariance matrix  is estimated in the kernel space and incorporated into the dual optimization problem of OSVM. Thus, we still have a convex optimization problem with a unique global solution, that can be reached easily using numerical methods. However, there are still some difficulties associated with COSVM application in real case problems, where data are sequentially obtained and learning has to be done from the first data. Besides, COSVM requires large memory and enormous amount of training time, especially for large dataset. 

Implementations for the existing One-Class Classification methods assume that all the data are provided in advance, and learning process is carried out in the same step. Hence, these techniques are referred to as batch learning. Because of this limitation, batch techniques show a serious performance degradation in real-word applications when data are not available from the very beginning. For such situation, a new learning strategy is required. 
\noindent Opposed to batch learning, incremental learning \cite{IncLearn} is more effective when dealing with non-stationary or very large amount of data. Thus, it finds its application in a great variety of situations, such as, visual tracking, software project estimation, brain computer interfacing, surveillance systems, etc... 

\section{Related Works}

Several learning algorithms have been studied and modified to incremental procedures, able to learn through time. Cauwenberghs and Poggio \cite{incdecsvm} proposed an online learning algorithm of Support Vector Machine (SVM). Their algorithm changes the coefficient of original Support Vectors (SV), and retains the Karuch-Kuhn-Tucker (KKT) conditions on all previously training data as a new sample acquired. Their approach have been extended by Laskov et al.\cite{IncSVM} to One-class SVM. However, the performance evaluation was only based on multi-class SVM. From their side, Manuel Davy et al. introduced in \cite{iOSVM} an online SVM for abnormal events detection (iOSVM). They proposed a strategy to perform abnormality detection over various signals by extracting relevant features from the considered signal and detecting novelty, using an incremental procedure. Incremental SVDD (iSVDD) proposed in \cite{iSVDD2} is also based on the control of the variation of the KKT conditions as new samples are added. An other approach to improve the classification performance is introduced in \cite{WOCSVM}. Incremental Weighted One-Class Support Vector Machine (WOCSVM) is an extension of incremental OSVM. The proposed algorithm aims to assign weights to each object of the training set, then it controls its influence on the shape of the decision boundary. An other incremental approach was introduced in \cite{Mahalinc}. The proposed algorithm is based on the Mahalanobis distance which takes into account the covariance in each feature direction and the different scaling of the coordinate axes. However, the aforementioned method involves inversion of the covariance matrix, which introduces complex computational problem in case of high dimensional datasets.  

\section{Aims and Objectives}

The proposed Incremental models based on Support Vector Machine (SVM) inherit the problem of classic SVM method which uses only boundary points to build a model, regardless of the spread of the remaining data. Also, none of them emphasizes the low variance direction, which results in performance degradation. Therefore, we solve this problem by introducing an incremental version of the COSVM approach, denoted by iCOSVM. In fact, iCOSVM incrementally emphasizes the low variance direction to improve classification performance. Our preposition aims to take advantages from the accuracy of COSVM procedure and we prove that it is a good candidate for learning in non-stationary environments.
The main goal of this PhD is to define a new classification framework able to deal with large learning dataset and  with  abilities to adjust itself according to the variability of the incoming data inside a stream. The key of our method is to construct a solution recursively, by adding one point at a time, and retain the Karush-Kuhn-Tucker (KKT) conditions on all previously acquired data.

\section{Methodology and Evaluation Strategy}

A deep and global bibliography study is essential to understand the basis of both batch and incremental approaches. A preliminary mathematical formulation of iCOSVM was performed.

In order to evaluate our method, we used both artiﬁcially generated datasets and real world datasets. For the experiments on artiﬁcially generated data, a number of 2D two-class data drawn from different set of distributions was generated.

For the real world case, datasets are collected from the UCI machine learning repository \cite{UCI} and picked carefully, so that we have a variety of sizes and dimensions, and we can, then, test the robustness of our iCOSVM.

To test the effectiveness of our proposed algorithm, it is appropriate to have two experimental scenarios. In the first scenario, we compare the iCOSVM to the COSVM and other contemporary batch algorithms to tease out the advantages of the incremental version, in terms of computational complexity and classification accuracy, over different batch learning algorithms. In the second scenario, we show the superiority of our iCOSVM over other well known incremental models, such as, iSVDD, iOSVM, WOCSVM, since it has the advantage of projecting data incrementally in low variance direction.

In future work, we aim to apply our algorithm on documents classification. In fact, the processing of documents starts with few data, whereas, after a while, many of them could be available and a huge amount of data that may carry valuable hidden information is produced. Therefore analytical tools have to take into account that most of data in modern systems arrives continuously. To carry out such experiments we can use Reuters dataset from the UCI machine learning repository.

For the implementation, we used specific toolbox in Matlab.


\begin{thebibliography}{4}

\bibitem{OCC}Juszczak, P.: Learning to recognize. A study on one-class classification and active learning. Technical University of Delft, The Netherlands (2006)

\bibitem{ConceptLearning}TAX, D.M.J.: One-class classification. Concept-learning in the absence of counter-examples. Technichal University of Delft (2001)

\bibitem{featpca}Tax, D.M.J. and M\"{u}ller, K.R.:Feature Extraction for One-Class Classification. In: Artificial Neural Networks and Neural Information Processing, pp.342-349. IEEE Press (2003)

\bibitem{COSVM}Khan, N.M., Ksantini, R., Ahmad, I.S. and Guan, L.:Covariance-guided One-Class Support Vector Machine. Pattern Recognition 47, 2165--2177 (2014)

\bibitem{IncLearn}Prachi Joshi and Dr.Parag Kulkarni: Incremental Learning: Areas and Methods - A Survey. International Journal of Data Mining \& Knowledge Management Process (IJDKP) 2, 43--51 (2012) 

\bibitem{incdecsvm}Cauwenberghs, G. and Poggio, T.: Incremental and Decremental Support Vector Machine Learning. In: Neural Information Processing Systems, pp.409-415. NIPS (2000) 

\bibitem{IncSVM}Laskov, P., Gehl, C. and Kr\"{u}ger, S.: Incremental Support Vector Learning: Analysis, Implementation and Applications. Journal of Machine Learning Research 7, 1909--1936 (2006)

\bibitem{iOSVM}Davy, M., Desorby, F., Gretton, A. and Doncarli, C.,: An Online Support Vector Machine for Abnormal Events Detection. Signal Processing, 2009--2025 (2005)

\bibitem{iSVDD2}Hua, X. and Ding, S.: Incremental Learning Algorithm for Support Vector Data Description. Journal of Software 6, 1166--1173 (2011)

\bibitem{WOCSVM}Krawczyk, B. and Wo\`{z}niak, M.: Incremental weighted one-class classiﬁer for mining stationary data streams. Journal of Computational Science 9 19--25 (2015) 

\bibitem{Mahalinc} Nader, P., Honeine, P. and Beauseroy, P.: Online One-class Classification for Intrusion Detection Based on the Mahalanobis Distance. In ESANN 2015 Proceedings, pp.567--572, Bruges,Belgium (2015) 

\bibitem{UCI}Lichman, M.: Machine Learning Repository, University of California, Iverine, School of Information and computer Sciences (2013) 

\end{thebibliography}
\end{document}